\pdfoutput=1

\documentclass[11pt]{article}

\usepackage{acl}

\usepackage{times}
\usepackage{latexsym}

\usepackage[T1]{fontenc}

\usepackage[utf8]{inputenc}

\usepackage{microtype}

\usepackage{graphicx}
\usepackage{booktabs}
\usepackage{subcaption}
\usepackage{hyperref}
\usepackage[capitalize]{cleveref}
\usepackage{multirow}
\usepackage{url}

\title{Judge a Sentence by Its Content to Generate Grammatical Errors}

\author{Chowdhury Rafeed Rahman \\
  National University of Singapore \\
  \texttt{rafeed@cse.uiu.ac.bd}}

\begin{document}
\maketitle
\begin{abstract}
Data sparsity is a well-known problem for grammatical error correction (GEC). Generating synthetic training data is one widely proposed solution to this problem, and has allowed models to achieve state-of-the-art (SOTA) performance in recent years. However, these methods often generate unrealistic errors, or aim to generate sentences with only one error. We propose a learning based two stage method for synthetic data generation for GEC that relaxes this constraint on sentences containing only one error. Errors are generated in accordance with sentence merit. We show that a GEC model trained on our synthetically generated corpus outperforms models trained on synthetic data from prior work.
\end{abstract}

\section{Introduction}

Grammatical error correction (GEC) is a well-known problem in NLP, and likely one of the most well-known to the general public as well---thanks to autocorrect services like Grammarly\footnote{\href{grammarly.com}{grammarly.com}}. Unfortunately, many current approaches to GEC rely on the presence of a large amount of parallel data for training the model, which can be very hard to find---especially for languages other than English. While correct sentences and incorrect sentences both exist in abundance, finding them in parallel is very challenging. This data sparsity problem has sparked research in generating synthetic corpora for GEC, which has been very successful in increasing performance for GEC~\cite{error3,error2,stahlberg2021synthetic}.

However, there is still room for improvement. Synthetic corpora often contain errors that are not human-like. This happens especially when error sentence is generated through back-translation \citep{error3} or random perturbations \citep{error2}. Recent research in this domain has attempted in generating realistic sentences tagged with only one error type per sentence~\cite{stahlberg2021synthetic}. Error type to be induced in a correct sentence is decided randomly, not based on the merit of the sentence. 

In this paper, we relax the requirement that sentences be tagged with only one error and develop a new method for synthetic error generation. Given a correct sentence, our error tagger model first judges what type of errors to induce in the sentence. Our corruption model then generates noisy version of that sentence based on those error types. For example, in a sentence with a principal verb and a countable noun (Example: \textit{He has bought many shoes.}), one would expect the generation of subject-verb agreement (\textit{has} transformed to \textit{have}) and/ or singular/ plural type error (\textit{shoes} transformed to \textit{shoe}). Our approach strives to achieve this feat.

\begin{figure}[htb!]
    \centering
    \begin{subfigure}[h]{0.60\textwidth}
        \centering
        \includegraphics[width=\textwidth]{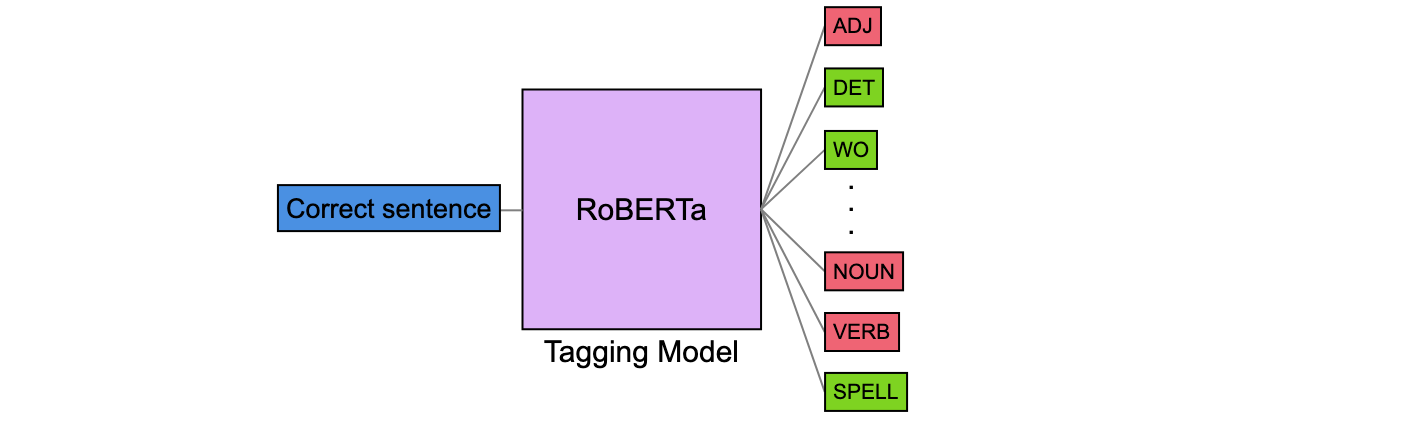}
        \caption{Error tagging model diagram.}
        \label{fig:tagging}
    \end{subfigure}
    \begin{subfigure}[h]{0.60\textwidth}
        \centering
        \includegraphics[width=\textwidth]{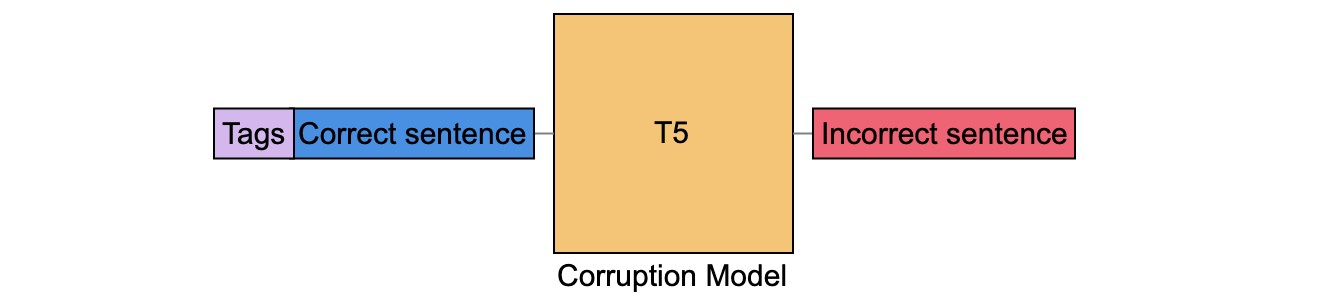}
        \caption{Corruption model diagram.}
        \label{fig:corruption}
    \end{subfigure}
    \caption{Diagram of our model. First, a correct sentence is assigned tags by the tagging model, based on a set of 24 possible error tags produced by ERRANT. Then this tag is appended to the correct sentence and our corruption model produces an incorrect sentence with these errors.}
    \label{fig:model}
\end{figure}

Our contributions in this paper are as follows:
\begin{enumerate}
    \item We propose a simple new method for synthetic data generation using the two-stage tagging and corruption pipeline shown in~\cref{fig:model}.
    \item We release a 700k sentence corpus generated using our method.
    \item Using this corpus, we show that a simple T5-small-based GEC model is able to outperform a model trained on an equal-sized subset of~\citeauthor{stahlberg2021synthetic}'s $C4_{200M}$ dataset.
\end{enumerate}

\section{Prior Work}
\citet{error3} used a back translation method along with a transformer sequence-to-sequence model for the first time to create noisy data from a clean corpus utilizing beam search from simple perturbation based approaches. This idea was previously proposed by \citet{error4} for machine translation.
\citet{error1} performed a detailed investigation of choices for pseudo-data generation in order to provide more training data to GEC models. Their experiments included two different versions (noisy and sampling) of back translation and a direct noise induction method. 
Later on, \citet{error2} proposed a simple unsupervised synthetic grammatical error generation procedure based on confusion sets. This confusion set was extracted with the help of an established spell checker. This spelling error dependent error corpus was used to train a sequence-to-sequence transformer model.
Recently, \citet{stahlberg2021synthetic} introduced error tag based synthetic grammatical error sentence generation. They experimented with various methods for matching the error tag distribution of a dev set. Furthermore, instead of using sequence to sequence transformer, their transformer model predicted possible edits for transforming a correct sentence into an incorrect one given an error tag. Using this method, they generated and released a large synthetic dataset, $C4_{200M}$, consisting of a 200M sentence subset of the $C4$ dataset~\cite{T5}. Using a Seq2Edits~\cite{stahlberg2020seq2edits} model trained on this data, they were able to achieve state-of-the-art (SOTA) performance.

\section{Methods}\label{sec:methods}
As shown in \cref{fig:model}, our generation process relies on two models: an error tagger and a corruption model\footnote{To use the same terminology as~\citeauthor{stahlberg2021synthetic}}. Given a correct sentence, our error tagging model assigns it a label consisting of a subset of the 24 ERRANT~\cite{bryant-etal-2017-automatic,felice-etal-2016-automatic} error tags described in~\cite{errant}. The concatenation of this label and the correct sentence is given to our corruption model which generates an incorrect version of the sentence following the errors in the tag.

\subsection{Error Tagging}
For the error tagging model, we use a pre-trained RoBERTa-base~\cite{liu2019roberta} from HuggingFace~\cite{wolf-etal-2020-transformers} and fine-tune it using a multi-label classification objective. That means multiple error tags can be predicted for a given input sentence simultaneously. This can be achieved by using Sigmoid activation function in each output layer node along with binary crossentropy loss for each of these node outputs. We use the correct sentences from the Lang-8~\cite{tajiri2012tense} training set as our training data. Because Lang-8 is a parallel corpus, we are able to generate ground truth labels using the error tagging tool, ERRANT~\cite{errant} to compare the correct and incorrect sentences. We disregard any sentences that receive a \texttt{noop} tag from ERRANT, as these are sentences that have no incorrect version, and we found that having the signal from these sentences significantly reduced performance when generating corrupt sentences in the next stage. Removing these examples reduces the size of Lang-8 from about 1M sentence pairs to 600k, but significantly improves the prediction quality of the model. 

\begin{figure*}[htb!]
    \centering
    \includegraphics[width=\textwidth]{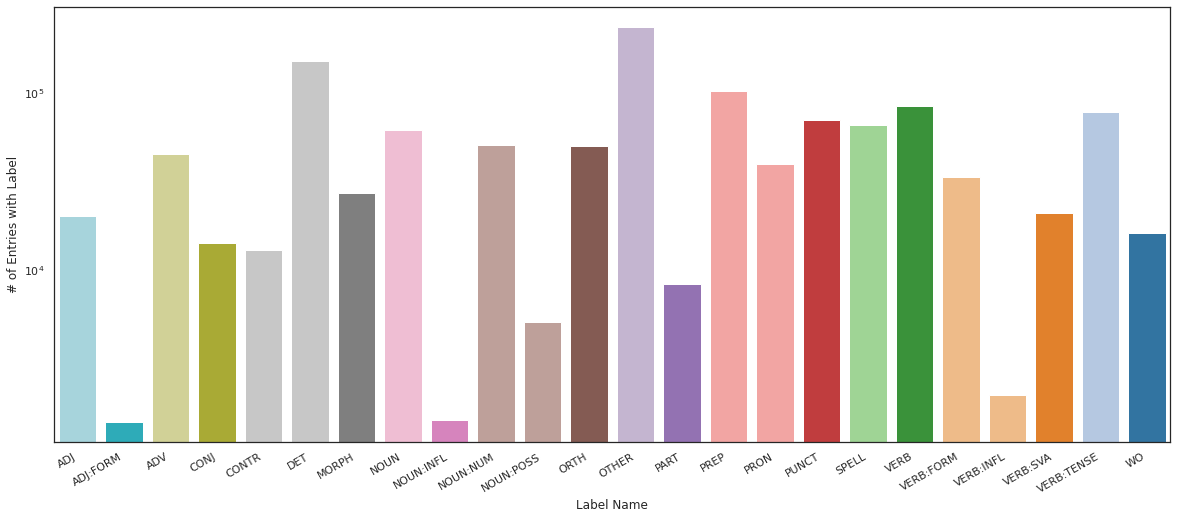}
    \caption{Label imbalance in Lang-8 data. Some labels are very common, occurring in over half the data, while other almost never occur, leading to large label imbalances.}
    \label{fig:imbalance}
\end{figure*}

One challenge in framing this as a multi-label classification problem where one sentence can, and often should, be tagged with multiple errors is that the labels in Lang-8 are highly imbalanced (as shown in~\cref{fig:imbalance}). Without correction, this can result in the model learning to simply never predict infrequent labels in favor of more frequent ones. To combat this, we use the oversampling method described in~\cite{charte2015addressing}, which corrects some of this imbalance without making the label distribution unrealistic. After oversampling, our training data is roughly 660k sentences. We hold out $10\%$ of this as our dev set. Our final model is trained for three epochs using cross-entropy loss and a learning rate of $1e^{-5}$ and a per-device batch size of 16 samples. To further help with the class-imbalance,  at inference time we use a threshold of 0.3 for predicting positive samples rather than the default 0.5 for our output layer nodes. We choose this threshold by searching for a threshold that optimizes the $F_{0.5}$ score of the model on Lang-8 validation portion. 

\subsection{Corruption Model}


We use a pretrained version of T5-base~\cite{T5} from the \textit{Happy Transformer} package built on top of HuggingFace’s transformer library. Both the input and the output of T5 model are text tokens irrespective of the type of task. \cref{fig:corruption} shows how we use T5 to generate corrupted sentence from grammatically correct sentence. \cref{tab:T5_Example} illustrates the input and output of two training samples provided to T5.

\begin{table*}[htb!]
\renewcommand{\arraystretch}{1.25}
\begin{tabular}{p{0.45\textwidth}p{0.25\textwidth}p{0.3\textwidth}}
\toprule
\textbf{Tagged Correct Sentence} & \textbf{Predicted Error Tags} & \textbf{Corrupt Version} \\ 
\midrule
grammar\_error: (aaaaaaaaabaabbaaaaaaaaaa) And he took in my favorite subjects like soccer . & \texttt{\textbf{\textcolor{red}{NOUN:NUM}}, OTHER, PART} & And he took in my favorite \textbf{\textcolor{red}{subject}} like soccer . \\
grammar\_error: (aaaaaaaaaababaabaaaaaaaa) His Kanji ability is much better than mine . & \texttt{\textbf{\textcolor{red}{NOUN:POSS}}, OTHER, \textbf{\textcolor{teal}{PRON}}} & His Kanji \textbf{\textcolor{red}{'s}} ability is much better than \textcolor{teal}{me} .    \\ \bottomrule
\end{tabular}
\caption{Example training samples provided to our T5 corruption model. The predicted error tags are generated by our error tagging model. Highlighted/bolded error tags are errors that are actually introduced by the corruption model and are highlighted in the corresponding color in the corrupt sentence.}
\label{tab:T5_Example}
\end{table*}

\textbf{Input:} We begin each input string with \textit{grammar\_error} as task conditioning for T5. This is followed by a bracketed 24 character string consisting of a's and b's representing each of the 24 types of errors. These tags are generated by the error tagger described in our previous subsection. \textit{b} in position \textit{i} means the presence of error type \textit{i} in the parallel corrupted sentence. The case for \textit{a} is just the opposite. This allows us to represent and generate sentences with multiple errors as shown in~\cref{tab:T5_Example}. Finally, after these tags, we append the grammatically correct sentence that we want to corrupt. 

\textbf{Output:} The output is simply the corrupted sentence produced by the model. \\


\begin{table*}[hbt!]
\begin{tabular}{|c|ccc|ccc|c|}
\hline
\multirow{2}{*}{\textbf{\begin{tabular}[c]{@{}c@{}}Model \\ Training Data\end{tabular}}} & \multicolumn{3}{c|}{\textbf{CoNLL-14 Test}}                                                         & \multicolumn{3}{c|}{\textbf{BEA-2019 Dev}}                                                          & \textbf{JFLEG Test}     \\ \cline{2-8} 
                                                                                         & \multicolumn{1}{c|}{P}              & \multicolumn{1}{c|}{R}              & F0.5                    & \multicolumn{1}{c|}{P}              & \multicolumn{1}{c|}{R}              & F0.5                    & BLEU                    \\ \hline
\textbf{$C4_{700K}$ (prev)}                                                                 & \multicolumn{1}{c|}{\textbf{50.25}} & \multicolumn{1}{c|}{14.72}          & \textit{33.89}          & \multicolumn{1}{c|}{33.47}          & \multicolumn{1}{c|}{14.18}          & \textit{26.31}          & \textit{83.26}          \\ \hline
\textbf{$C4_{700K}$ (new)}                                                                  & \multicolumn{1}{c|}{45.73}          & \multicolumn{1}{c|}{16.96}          & \textit{34.15}          & \multicolumn{1}{c|}{40.72}          & \multicolumn{1}{c|}{14.93}          & 30.29                   & 82.70                   \\ \hline
\textbf{Lang8 + FCE}                                                                     & \multicolumn{1}{c|}{46.01}          & \multicolumn{1}{c|}{25.66}          & \textit{39.71}          & \multicolumn{1}{c|}{48.79}          & \multicolumn{1}{c|}{31.64}          & \textit{44.02}          & \textit{84.48}          \\ \hline
$C4_{700k}$~\textbf{(new) + Lang8 + FCE}                                                          & \multicolumn{1}{c|}{46.15}          & \multicolumn{1}{c|}{\textbf{25.84}} & \textit{\textbf{39.88}} & \multicolumn{1}{c|}{\textbf{49.24}} & \multicolumn{1}{c|}{\textbf{32.53}} & \textit{\textbf{44.65}} & \textit{\textbf{85.13}} \\ \hline
\end{tabular}
\caption{Performance of a T5-small GEC model trained on different datasets and evaluated on CoNLL-2014, BEA-19 (dev), and JFLEG (test). Best results are in bold.}
\label{tab:GEC-performance}
\end{table*}


\subsection{Data Generation}
After training the tagging and corruption models, we use this pipeline to generate a \~700k sentence synthetic dataset using a subset of the C4 dataset~\cite{T5}. To do this, we first obtain 700k grammatically correct sentences from C4 and tokenize them using SpaCy~\cite{spacy}\footnote{We do this because our model is trained exclusively on tokenized data.} before passing them through our tagging model. We use these tags to generate the prefix described in~\cref{sec:methods} and shown in~\cref{tab:T5_Example}, and finally generate our corpus using our corruption model as described in the previous subsection. Our T5 corruption model is auto-regressive. In each step of token generation, we probabilistically sample from the top 50 token predictions. We have experimented with both beam search and different number of top predictions. But the mentioned setting gives the most realistic errors. The resulting corpus is available here: \textit{{https://bit.ly/parallel-c4}}.

\section{Results}
\label{sec:results}
\subsection{Error Tagging}
We evaluate our error tagging model on the Lang-8 validation dataset, which we pre-process in the same way as the train dataset, by removing all sentence pairs that get \texttt{noop} ERRANT tags. This results in 6k test samples. The precision, recall and F0.5 score achieved by our model are 39.49, 42.27 and 40.02, respectively.

\subsection{GEC Performance Using Synthetic Corpus}

We use T5-small model to test what kind of  performance gain can be seen from using our synthetically generated data to train a GEC model. We evaluate on CoNLL-2014~\cite{ng2014conll}, the BEA-19 development set~\citep{bryant2019bea}, and the JFLEG test set~\citep{napoles2017jfleg}, and report our results in ~\cref{tab:GEC-performance}. 

To train our models, we generate 700K GEC samples using 700K randomly sampled correct sentence from C4~\citep{T5}. We refer to this data as $C4_{700k}$ ~(new). We train a T5-small model on our generated data and report the results (see $C4_{700K}$ (new) of Table \ref{tab:GEC-performance}). For comparison, we also sample 700k synthetic GEC samples from~\citeauthor{stahlberg2021synthetic}'s $C4_{200M}$ (this is the latest research on GEC synthetic error data generation) and train another model only on these samples (see $C4_{700K}$ (prev) of Table \ref{tab:GEC-performance}). The model trained on our data outperforms the previous C4 trained model on both CoNLL-14 and BEA-2019, with the performance difference for BEA being especially large. However, on JFLEG, our approach underperforms slightly. $C4_{700K}$ (prev) was generated using a large version of Transformer model while we use a base version model. Our method still does better which holds significant value.\\
We also analyze the impact of adding $C4_{700k}$ ~(new) data in training along with Lang-8~\citep{mizumoto2011mining} and FCE~\citep{yannakoudakis2011new} dataset. The last two rows of the table show the improvement of performance in all cases after the addition of our proposed synthetic dataset. 

\section{Conclusion}
Our method for synthetic error generation was able to improve upon the state of the art in synthetic error generation, which we believe yields several useful insights for future work. First, framing the tagging step as a multi-label rather than multi-class classification problem seems to be an important distinction.The examples generated by our model seem relatively realistic. This work does not outperform any SOTA GEC systems. Rather, this research is performed using relatively small models and relatively small synthetic error dataset as a proof of concept. Future work should perform extensive experiments based on the proposed method to achieve SOTA performance in GEC.  
Additional methods to correct for imbalance in the error tagging training data should be explored, as this is one of the biggest barriers in performance for the tagging model. Code and dataset of this research are available at: \textit{https://bit.ly/3Kc0oY8}.

\clearpage


\bibliography{main}
\bibliographystyle{acl_natbib}

\end{document}